\documentclass{article}

\usepackage[margin=1in]{geometry}
\usepackage[utf8]{inputenc}
\usepackage[preprint]{global_aliases}
\usepackage{local_aliases}
\usepackage{amsmath}

\usepackage{balance} 
\usepackage{amsthm}
\usepackage{hyperref}

\title{Strategic Arms With Side Communication prevail over Low-Regret MAB Algorithms}

\threeauthors
 {Ahmed Ben Yahmed\sthanks{ENS Paris Saclay, France \href{mailto:ahmed.ben\_yahmed@ens-paris-saclay.fr}{(ahmed.ben\_yahmed@ens-paris-saclay.fr)}}}
	{}
 {Clément Calauzènes\sthanks{CRITEO AI lab, Paris, France \href{mailto:c.calauzenes@criteo.com}{(c.calauzenes@criteo.com)}}}
	{}
{Vianney Perchet\sthanks{ CREST, ENSAE, Palaiseau, France /
          CRITEO AI Lab, Paris, France \href{mailto:vianney.perchet@normalesup.org}{(vianney.perchet@normalesup.org)}}}
	{}

\date{\today}

 \usepackage{lipsum}
 \usepackage{fancyhdr}
\fancypagestyle{firstpage}{\fancyfoot[L]{ \scriptsize \copyright\ Copyright 2024 IEEE. Published in ICASSP 2024 - 2024 IEEE International Conference on Acoustics, Speech and Signal Processing (ICASSP), scheduled for 14-19 April 2024 in Seoul, Korea. Personal use of this material is permitted. However, permission to reprint/republish this material for advertising or promotional purposes or for creating new collective works for resale or redistribution to servers or lists, or to reuse any copyrighted component of this work in other works, must be obtained from the IEEE. Contact: Manager, Copyrights and Permissions / IEEE Service Center / 445 Hoes Lane / P.O. Box 1331 / Piscataway, NJ 08855-1331, USA. Telephone: + Intl. 908-562-3966.}}

\begin{document}

\ninept
\maketitle
\begin{abstract}

    In the strategic multi-armed bandit setting, when arms possess perfect information about the player's behavior, they can establish an equilibrium where: {\bf 1.} they retain almost all of their value, {\bf 2.} they leave the player with a substantial (linear) regret. This study illustrates that, even if complete information is not publicly available to all arms but is shared among them, it is possible to achieve a similar equilibrium. The primary challenge lies in designing a communication protocol that incentivizes the arms to communicate truthfully.\\

\end{abstract}

\vspace{-3mm}
\begin{keywords}
    multi-armed bandit, strategic arms, communication through a network, Nash equilibrium.
\end{keywords}

\thispagestyle{firstpage}
\section{Introduction}

The concept of the strategic multi-armed bandit extends the traditional multi-armed bandit (MAB) problem by incorporating the utility aspect of the arms. In this context, arms have the ability to report values that differ from the observed rewards. Formally, we consider a set of $K$ stochastic arms, each arm $k$ is characterized by its own reward distribution $D_k$ with a mean denoted as $\mu_k = \mathbb{E}[D_k]$. To maintain clarity while broadening our perspective, we assume an order such that $1 \geq \mu_1 \geq \mu_2 \geq \cdots \geq \mu_K \geq 0$.
During each round $t$, the player pulls an arm $k_t$. Subsequently, the chosen arm observes a reward $r_{k_t, t} \sim D_{k_t}$, and it has the possibility to report a value $x_{k,t} \neq r_{k,t}$ to the player while retaining $r_{k,t} - x_{k,t}$ as its own utility. Importantly, only arm $k_t$ possesses knowledge of the actual observed reward $r_{k_t,t}$, whereas the player is only aware of $x_{k_t,t}$ and remains unaware of the withheld portion. Therefore, the player's decision is based on the information collected up to time $t$, which can be formally encoded in the filtration $\mathcal{F}_{P,t} = \{k_1, x_{k_1, 1}, \ldots, k_t, x_{k_t, t}\}$. We define $x_{k}$ (and $r_{k}$) as the concatenation of reported values (and rewards, respectively) over $T$ rounds. Accordingly, the utility associated with arm $k$ can be expressed as:
\begin{align}
\mathcal{U}_{k}(x_{k},x_{-k}) = \mathbb{E} \left[\sum_{t=1}^{T} (r_{k_{t},t} - x_{k_{t},t}) \cdot \indicator{k_{t}=k}\right]
\end{align}
This strategic scenario introduces a game-like dynamic that engenders a competition of objectives between the player, who strives to minimize regret (see section~\ref{section: regret}), and the arms, which are driven by the pursuit of maximizing their utilities. This model encapsulates a diverse array of dynamic agency dilemmas wherein the player selects an arm (agent) to execute a task on his behalf, and the associated cost remains concealed from the player due to his limited domain or market knowledge. Broadly speaking, this model can be viewed as an extension to the multi-agent realm, akin to the principal-agent problem in contract theory~\cite{principal-agentProblem,principal-agentProblem1}, albeit with multiple agents in play. It significantly extends the standard MAB problem, as arms can utilize this reporting mechanism to influence the player's decisions. For example, arms may opt to report higher values initially to increase their chances of being selected in later rounds. Conversely, they may report lower values at the outset to decrease the reserve price in auctions~\cite{amin}. Furthermore, our study takes into consideration the existence of side communications among arms, governed by predefined rules. This consideration is motivated by real-world scenarios in which such interactions are prevalent and influential.

\subsection{Related Work}
Previous studies, such as \cite{SideOb, SideOb1, SideOb2}, have already examined scenarios involving connected and communicating arms. In these scenarios, pulling an arm $k$ at time step $t$ not only provides information about arm $k$ itself but also reveals information about some related arms. A typical example of such a situation is advertising on social networks, where a decision-maker targets individual users of an online platform with promotions, hoping to maximize purchases. However, in this context, the arms (i.e users) are connected and capable of communication. Conceptually, pulling an arm $k$ triggers instantaneous communication through the arms, revealing aggregated information about all arms related to $k$ and itself \cite{SideOb}. The concept of a strategic arm in the MAB setting was first introduced in the groundbreaking work by \cite{pmlr-v99-braverman19b}. This notion highlights the challenge of dealing with arms that are not limited to providing their true rewards but can instead manipulate the player to maximize their own utilities. To achieve this, authors present a equilibrium strategy for such arms especially when $\mu_1-\mu_2 \leq \frac{\mu_1}{K}$, enabling them to leave the player with only a minimal reward. This strategic approach poses a significant obstacle for any low-regret algorithm employed by the player.

\subsection{Contributions}
This study addresses the limitation of \cite{pmlr-v99-braverman19b} where arms need full information to be able to have an equilibrium strategy in which they can extract (almost) all the value. The new strategy introduced here, which includes a proper communication protocol, enables them to achieve equilibrium for full surplus extraction and prevents the player from generating high revenue, regardless of the low-regret MAB algorithm the player chooses. We build upon the strategy presented in \cite{pmlr-v99-braverman19b}, which serves as a foundation for our work. The main challenge in constructing an equilibrium strategy that doesn't rely on public knowledge of the entire history is to guarantee that the communication scheme doesn't encourage arms to convey false information. Indeed, from a game-theoretic perspective, both the report $x_{k_t, t}$ to the player and the information they share with other arms are components of strategic behavior. We support our claims with theoretical analysis and experimental results.

\section{Modeling Communication}  
\begin{figure}[htb]
     \vspace{-1mm}
     \centering
     \includegraphics[width=0.5\linewidth]{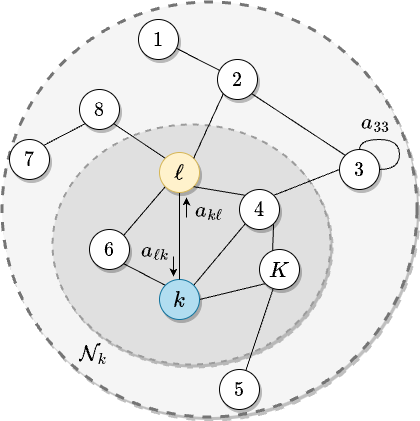}
     \vspace{-3mm}
     \caption{ 
     Arms are linked by a graph topology and can share information over their shared edges. The neighborhood of an arm is the collection of all arms linked to it. The neighborhood of arm k is marked by the highlighted area and denoted by $\mathcal N_k$. We assign a pair of nonnegative scaling weights, \{\( a_{k \ell} \),\( a_{\ell k} \)\}, to the edge connecting $k$ and $\ell$. The scalar \( a_{\ell k} \) will be used by arm $k$ to scale information it receives from arm $\ell$; this scaling can be interpreted as a measure of confidence~\cite{profbook1}.\label{Fig:graph}}
     \vspace{-3mm}
\end{figure}

We assume that the arms are interconnected and form a network modeled by a graph. This setting is widely used in distributed learning and multi-agent communication \cite{ProfBook} as it provides a  stronger privacy protection and reduced risk of communication bottleneck. Arms will be linked by some graph topology (see Fig.~\ref{Fig:graph}) that allows neighboring arms to share information. At any point in time, no single arm will have access to all the information that is available across the graph. Additionally, all types of message passing occur simultaneously throughout the graph. This means that all information at all arms is updated instantaneously and equivalently at the same time (i.e the network has a synchronized clock).

\section{low-regret MAB algorithms}
\label{section: regret}
 
The efficiency of MAB algorithms is most commonly compared with the notion of regret. Regret measures the cumulative loss incurred over $T$ rounds by choosing arm $k_t$ at time $t$ instead of the best arm $k^\star$. Ideally, we aim for a low-regret, which implies convergence to the best arm, at least asymptotically. Formally, let $k_t$ be the strategic arm selected by the player at round $t$ using algorithm $A$. The selected arm observes a reward $r_{k_t, t}$ that we consider in this work in $[0,1]$. Then, it reports , in an adversarial fashion, a value $x_{k_t, t}$. The regret of the algorithm $A$ is the random variable:
\begin{align}
    \mathcal{R}(A)=\max_k \sum_{t=1}^{T} x_{k,t} - \sum_{t=1}^{T} x_{k_t,t}
\end{align}
We follow the $(\rho,\delta)$-low-regret definition given in \cite{pmlr-v99-braverman19b}, i.e an algorithm $A$ is a $(\rho,\delta)$-low-regret for the MAB problem if with probability $1-\rho$,
\begin{align}
    \mathcal{R}(A) \leq \delta
\end{align}
The majority of MAB algorithms, particularly in the adversarial scenario -- which is relevant here, considering that the reporting protocol is widely regarded as an adversarial setting -- assign a probability $p_{k,t}$ to each arm $k \in \{1, \cdots, K\}$ to be pulled at round $t$. The algorithm then selects the arm according to these probabilities. The probability $p_{k,t}$ is primarily determined by the parameters of the algorithm $A$, an intrinsic information $I_k$ of arm $k$, and the average value of this information $\tilde{I}$ across all the arms. Formally it can be seen as the following function:
\begin{align}
    \Pr: \{\text{set of parameters}\}\times \lR \times \lR &\rightarrow [0,1] \\
    (A,I_{k,t},\Tilde{I}_{t}) &\mapsto p_{k,t}
\end{align}
For instance, in the EXP3 algorithm \cite{AuerExp3}, to compute the probability $p_{k,t}$ of choosing arm $k$ at time $t$, we require the parameter $\gamma$, which determines the trade-off between exploration and exploitation. Additionally, we need the weighted estimated reward which is the intrinsic information $I_{k,t}$ associated with arm $k$ at time $t$, and the sum of these rewards over all arms, denoted as $K\tilde{I}_{t}$.
\section{Arms' Strategy That prevails over low-regret MAB Algorithms}
In this section, we extend the work presented in \cite{pmlr-v99-braverman19b} and introduce a strategy that enables strategic arms with restricted communication to reach an $\epsilon$-Nash equilibrium while providing only a marginal utility to the player. The Strategy~\ref{Strategy} adopts market sharing techniques, where arms select their actions in a way that ensures they are chosen an equal number of times. As a result, the player receives only minimal revenue, as he is unable to commit to selecting the best arm consistently, deviating from the traditional bandit setting. This strategy does not assume that the arms possess prior knowledge of their own distributions or the history of selected arms. In other words, their respective information available at time $t$ is less complex compared to the setting presented in \cite{pmlr-v99-braverman19b}. The strategy is presented as follow: let $A$ be the low-regret MAB algorithm used by the player
, $N_{k}(t)$ be the number of times arm $k$ has been pulled up to time $t$ 
and $\mathbb{A}=[a_{k\ell}]$ the combination matrix describing the communication graph topology. Set 
$B=7\sqrt{KT\delta}$ and 
$\theta= \sqrt{\frac{K\delta}{T}}$. Then the strategy that the arms shall use is Strategy~\ref{Strategy}.
\newcommand{\algrule}[1][.2pt]{\par\vskip.5\baselineskip\hrule height #1\par\vskip.5\baselineskip}

\newenvironment{strategy}[1][htb]
  {\renewcommand{\algorithmcfname}{Strategy}% Update algorithm name
   \begin{algorithm}[#1]%
  }{\end{algorithm}}

\begin{strategy}

    \DontPrintSemicolon
    \For{$t=1,\cdots,T$}{
    Update $I_{k,t}$  the intrinsic information available for all arm $k$ at time $t$. Initiate $\Tilde{I}_{k,0}=I_{k,t} , \forall k$.\\

    \For{$n=1,\cdots,\tau$}{
    \For{$  k=1,\cdots,K$}{
      $\Tilde{I}_{k,n}=\underset{\ell\in N_{k}}{\sum}a_{\ell k}I_{\ell,n-1}$} 
    }
    \For{$k=1,\cdots,K$}{
    \textbf{If} at any time $s\leq t$ in the past $ N_{k}(s) < \frac{s}{K}-B $ then arm $k$ defects and offers its full value $x_{k,t}=r_{k,t}$.\label{defctst}\\
    \textbf{Else} arm $k$ computes the probability $\hat{p}_{k,t}=\Pr(A,I_k,\Tilde{I}_{k,\tau})$ and offers $x_{k,t}=\theta(1-\hat{p}_{k,t})$. 
    }
   }
   \vspace{-1mm}
  \caption{Equilibrium strategy }\label{Strategy}
\end{strategy}
 \vspace{-1mm}
This strategy consists of two parts. In the first part, each arm updates its information individually. For example, if an arm is pulled, it updates its reward information based on the received reward. If it is not chosen, the information remains the same as in the previous time step $t-1$. Then, iteratively, the arms use the communication scheme to compute a local estimate, denoted as $\tilde{I}_{k,\tau}$, of the true average information $\tilde{I}_t$. During the second part, each arm $k$ uses its estimated average to compute the probability $\hat{p}_{k,t}$ of being selected by the player. The arm then adjusts its offer based on this probability. If $\hat{p}_{k,t}$ is high, the arm sets a low value for $x_{k,t}$, allowing other arms to have a higher chance of being chosen. On the other hand, if $\hat{p}_{k,t}$ is low, the arm sets a high value for $x_{k,t}$, increasing its own chance of being chosen since other arms are more likely to be selected. By following this approach, the arms achieve an equilibrium in terms of market sharing, which results in lower revenue for the player. The parameter $\theta$ in $x_{k,t}$ is introduced to facilitate the theoretical analysis of the strategy. In the next section, we will proceed with the theoretical analysis, where we will demonstrate that if the arms follow this strategy, they will reach an equilibrium, discouraging any defection.

\section{Theoretical analysis}
To facilitate the analysis, we suppose that $K\leq \frac{T^{\frac{1}{3}}}{\log(T)},\rho \leq \frac{1}{T^2}$ and $\delta \geq \sqrt{T\log(T)}$.

\subsection{Reliable approximation of pulling probabilities}

First, we will demonstrate the utility of the communication steps and how, after a sufficient number of iterations $\tau$, the local values $\tilde{I}_{k,\tau}$ serve as accurate estimates of the true average $\tilde{I}_t$ which allows each arm $k$ to compute an accurate estimate $\tilde{p}_{k,t}$ of $p_{k,t}$. To do so, we start by introducing some assumptions that are commonly used in the literature \cite{ProfBook,yuan2015convergence,profbook1,Kayaalp_2022}.

\begin{assumption}[\textbf{Doubly-stochastic combination matrix}]\label{dsm}
    The combination matrix $\mathbb{A}=[a_{\ell k}]$ representing the graph topology is doubly-stochastic and symmetric. This means that the matrix has non-negative elements and satisfies:
\begin{equation}
        \mathbb{A}\mathds{1}_K = \mathds{1}_K \footnotemark, \mathbb{A}^{\mathsf{T}}=\mathbb{A}
\end{equation}
\footnotetext{$\mathds{1}_K$ is a vector of length $K$ consisting of ones.}
We also assume that the matrix $\mathbb{A}$ is primitive. This implies that there exist paths, in both directions, between any two distinct nodes with nonzero scaling weights. Additionally, there is at least one non-trivial self-loop present, meaning that $a_{kk}>0$ for at least one node $k$.
\end{assumption}
By applying the \textit{Perron-Frobenius} theorem, Assumption~\ref{dsm} states that the mixing rate $\lambda$ of the combination matrix (i.e., the spectral radius of $\mathbb{A}-\frac{1}{K}\mathds{1}_{K} {\mathds{1}_{K}^\mathsf{T}}$) is strictly less than 1:
\begin{align}
\lambda < 1 \label{lamdbaCond}
\end{align}

\begin{assumption}[\textbf{Lipschitz mapping}]\label{lip}
The mapping $\Tilde{I}\mapsto \Pr(., ., \Tilde{I})$ is Lipschitz, namely: $\exists L \in \mathbb{R}_+$ such that $\forall \Tilde{I}, \Tilde{I}^\prime$:
\begin{align}
    || \Pr(., ., \Tilde{I}) -\Pr(., ., \Tilde{I}^\prime)|| \leq L || \Tilde{I} -\Tilde{I}^\prime||
\end{align}
\end{assumption}
Assumption~\ref{lip} is valid since we are considering a finite horizon $T$.

\begin{theorem}[\textbf{Network disagreement}]\label{NetDis}
    Under Assumption~\ref{dsm}, the network disagreement between the true average $\tilde{I}_t$ and the local estimates $\tilde{I}_{k,\tau}$ converges to zero.
    \begin{align}
        \frac{1}{K} \overset{K}{\underset{k=1}{\sum}} \norm{\tilde{I}_{k,\tau}-\Tilde{I}_t}^2 \leq \alpha_t \lambda^{2\tau} \underset{\tau\to +\infty}{\longrightarrow} 0  \; \text{ with } \alpha_t >0
    \end{align}
\end{theorem}

\underline{\textbf{Proof:}}
for generality we suppose that the dimension of variables is $M$, i.e $dim(I_{k,t})=dim(\tilde{I}_t)=dim(\tilde{I}_{k,t})=M$. We begin by defining the following variable that aggregates the local variables of each arm into a single variable:
\begin{align}
    \mathcal{\tilde{I}}_n &\triangleq \mathrm{col} \{\Tilde{I}_{1,n}, \cdots,\Tilde{I}_{k,n}\} \\
    \mathcal{A} &\triangleq \mathbb{A} \otimes \1 \footnotemark
\end{align}
\footnotetext{$\1$ refers to the identity matrix of dimension $M$, and $\otimes$ represents the Kronecker product.}
Where $\mathcal{\tilde{I}}_n$ is a vector of length $K \times M$, obtained by vertically concatenating vectors enclosed in brackets. 
We express the update in the communication scheme using a more concise notation:
\begin{align}
    \mathcal{\tilde{I}}_n &= \mathcal{A}^\top  \mathcal{\tilde{I}}_{n-1} \\
    \Rightarrow \left(\frac{1}{K} \Itop \otimes \1\right) \mathcal{\tilde{I}}_n & \overset{{(\footnotemark)}}{=} \left(\frac{1}{K} \Itop \otimes \1\right) \mathcal{\tilde{I}}_{n-1} 
\end{align}
\footnotetext{$\left(\frac{1}{K} \Itop \otimes \1\right)\mathcal{A}^\top=\frac{1}{K} \Itop \otimes \1$}
It should be noted that due to the nature of the combination matrix $\mathbb{A}$, the true average $\tilde{I}_t$ is equivalent to the average of the variables $\tilde{I}_{k,n}$ for any given $n$. So we write:
\begin{align}
    \tilde{I}_{t}= \frac{1}{K} \overset{K}{\underset{k=1}{\sum}} \tilde{I}_{k,n}= \left(\frac{1}{K} \Itop \otimes \1\right) \mathcal{\tilde{I}}_n
\end{align}
and we define the extended average as a vector of length $K \times M$:
\begin{align}
    \mathcal{\tilde{I}}_t \triangleq \I \otimes  \tilde{I}_{t}= \left(\frac{1}{K} \I\Itop \otimes \1\right) \mathcal{\tilde{I}}_n
\end{align}
We get:
\begin{align}
    &\mathcal{\tilde{I}}_n - \mathcal{\tilde{I}}_t =  \left(\mathcal{A}^\top-\frac{1}{K} \I\Itop \otimes \1\right) \mathcal{\tilde{I}}_{n-1} \\
    &\overset{{(\footnotemark)}}{=} \left(\mathcal{A}^\top-\frac{1}{K} \I\Itop \otimes \1\right) \left(\1-\frac{1}{K} \I\Itop \otimes \1\right)  \mathcal{\tilde{I}}_{n-1} \\
    &=\left(\mathcal{A}^\top-\frac{1}{K} \I\Itop \otimes \1\right) \left(  \mathcal{\tilde{I}}_{n-1} - \mathcal{\tilde{I}}_t \right)
\end{align}
\footnotetext{ $ \left(\mathcal{A}^\top-\frac{1}{K} \I\Itop \otimes \1\right) \left(\1-\frac{1}{K} \I\Itop \otimes \1\right)=\mathcal{A}^\top-\frac{1}{K} \I\Itop \otimes \1$}
Taking the square norm:
\begin{align}
    \norm{\mathcal{\tilde{I}}_n - \mathcal{\tilde{I}}_t }^2 &= \norm{\left(\mathcal{A}^\top-\frac{1}{K} \I\Itop \otimes \1\right) \left(  \mathcal{\tilde{I}}_{n-1} - \mathcal{\tilde{I}}_t \right)}^2 \\
    &\leq \lambda^2  \norm{ \mathcal{\tilde{I}}_{n-1} - \mathcal{\tilde{I}}_t }^2 
\end{align}
Iterating from $\tau$ to $0$:
\begin{align}
  \norm{\mathcal{\tilde{I}}_\tau - \mathcal{\tilde{I}}_t}^2 \leq \lambda^{2\tau} \norm{ \mathcal{\tilde{I}}_{0} - \mathcal{\tilde{I}}_t }^2
\end{align}
Taking $\alpha_t=\frac{\norm{ \mathcal{\tilde{I}}_{0} - \mathcal{\tilde{I}}_t }^2}{K}$ finishes the proof.

\begin{corollary}\label{corol1}
    Under Assumption~\ref{lip} and using Theorem~\ref{NetDis}, we have:
    \begin{align}
        |p_{k,t}-\hat{p}_{k,t}|&=|\Pr(A,I_{k,t},\tilde{I}_{t})- \Pr(A,I_{k,t},\tilde{I}_{k,t})| \\
        &\leq L \sqrt{\alpha_t} \lambda^\tau \underset{\tau \to +\infty}{\longrightarrow} 0
    \end{align}
\end{corollary}
Therefore, we have demonstrated that after a sufficient number of iterations $\tau$, $\hat{p}_{k,t}$ provides a reliable approximation for $p_{k,t}$. This allows the arms to calibrate their rewards as if they have access to the complete information available to the player.

\subsection{Equilibrium resulting from Strategy~\ref{Strategy}}
Following Strategy~\ref{Strategy}, arms won't defect and  will achieve a market sharing situation where each arm is pulled approximately an equal number of times. This renders the utilized low-regret MAB algorithm inefficient. This observation is formalized as follows:
\begin{restatable}[]{theorem}{theo}
\label{theoPrinc}

    If arms use Strategy~\ref{Strategy}, then with high probability $(1-\frac{3}{T}), N_{k}(t)\geq \frac{t}{K}-B, \forall t\in[T], k\in[K]$ and they will be in an $O(\sqrt{KT\delta})$-Nash equilibrium.

\end{restatable}
\underline{\textbf{Proof sketch\footnotemark:}} \footnotetext{The detailed proof is omitted due to space limitations.}if arms faithfully adhere to the Strategy~\ref{Strategy} denoted as $S^\star$, then by employing Corollary~\ref{corol1} and similar arguments as in~\cite{pmlr-v99-braverman19b}, we can demonstrate that with high probability (1 - $\frac{3}{T}$), $N_{k}(t) \geq \frac{t}{K} - B$ for all $t \in [T]$ and $k \in [K]$. This implies that arms do not defect, and step~\ref{defctst} of the strategy is never executed. To establish equilibrium, we will evaluate the utility of arm $k$ when all arms are adhering to $S^\star$, while the player employs a low-regret MAB algorithm. We show that:
\begin{align}
    \mathcal{U}_{k}( S^\star_{k},S^\star_{-k})
    & \geq \frac{\mu_{k}T}{K} - O(\sqrt{KT\delta})
\end{align}
On the other hand, if arm $k$ plays any strategy $S$ other than $S^\star$, we can demonstrate that:
\begin{align}
    \mathcal{U}_{k}(S_{k},S^\star_{-k}) &\leq  \frac{\mu_{k}T}{K} + O(\sqrt{KT\delta})
\end{align}
Therefore, for all strategies $S$ different from $S^\star$, we can derive the following inequality:
\begin{align}
     \mathcal{U}_{k}(S_{k},S^\star_{-k})- \mathcal{U}_{k}(S^\star_{k},S^\star_{-k}) \leq O(\sqrt{KT\delta})
\end{align}
showing that $(S^\star_1,\ldots,S^\star_K)$ is an $O(\sqrt{KT\delta})$-Nash equilibrium for all arms. Therefore, it becomes evident that the equilibrium is primarily determined by the number of times each arm is pulled. At equilibrium, we observe that each arm is pulled approximately the same number of times and receives in average $\frac{\mu_k}{K}$ per round. If arms choose to deviate from this strategy by dishonestly reporting either \emph{their values to the player} or \emph{the values communicated to their neighbors}, one of two scenarios will unfold.
In the first scenario, this deviation will not impact the number of times each arm is pulled, thus failing to activate the defection step~\ref{defctst}. Consequently, the utility of the arms remains unaffected, and the equilibrium remains intact.
In the second case, the defection step~\ref{defctst} is triggered, resulting in arm 1, which has the highest real mean value, emerging as the winner. In this case, it gains at maximum $\mu_1-\mu_2$ per round. However, it's important to note that $\mu_1-\mu_2 \leq \frac{\mu_1}{K}$ by assumption. Given this condition, there is no incentive for arms to deviate from the strategy, as the potential gain from defection is less than what they can achieve by adhering to the equilibrium strategy.\\
\\
This equilibrium proves detrimental to the player, resulting in constrained revenue, irrespective of the low-regret algorithm employed:
\vspace{-5mm}
\begin{corollary}\label{lowRev}
    If arms follow Strategy~\ref{Strategy}, the player gets at most $O(\sqrt{KT\delta})$ revenue.
\end{corollary}
\underline{\textbf{Proof:}}
given that playing according to Strategy~\ref{Strategy} implies that with high probability $\forall t\in[T],k\in[K], N_{k}(t) \geq \frac{t}{K} - B$, arms won't defect and the player gets $T\theta=O(\sqrt{KT\delta})$. In the case of the low probable event the player will get at most $T$. So the player revenue is:
\begin{align}
    \text{Player-revenue} &\leq (1-\frac{3}{T})O(\sqrt{KT\delta}) + \frac{3}{T}T\\
    \Rightarrow \text{Player-revenue}&\leq O(\sqrt{KT\delta})
\end{align}

\section{Experiment}
In this section we test Strategy~\ref{Strategy} against an adapted version of EXP3.P~\cite{bubeck2012regret} as it is done in~\cite{SideOb} to take into consideration the existence of the side communication. It is a $\left(\rho,O(\sqrt{T\log(K\rho^{-1})})\right)$-low-regret MAB algorithm. The intrinsic information for arm $k$ is the exponential of its weighted estimated cumulative gain. 
We create a random Erdos-Rényi graph over $K=10$ nodes, where each pair of nodes are linked independently with probability $p=0.6$. Arms are modeled as Bernoulli random variables. For most arms, mean is set at 0.4, while three specific arms have different means: 0.8, 0.85, and 0.9. The combination matrix $\mathbb{A}$ is generated using the Metropolis rule~\cite{metropolis}. We run the experiment for $T=5.10^5$ rounds and set $\tau=50$.

\begin{figure}[!htb]
     \vspace{-2mm}
     \centering
     \includegraphics[width=1\linewidth]{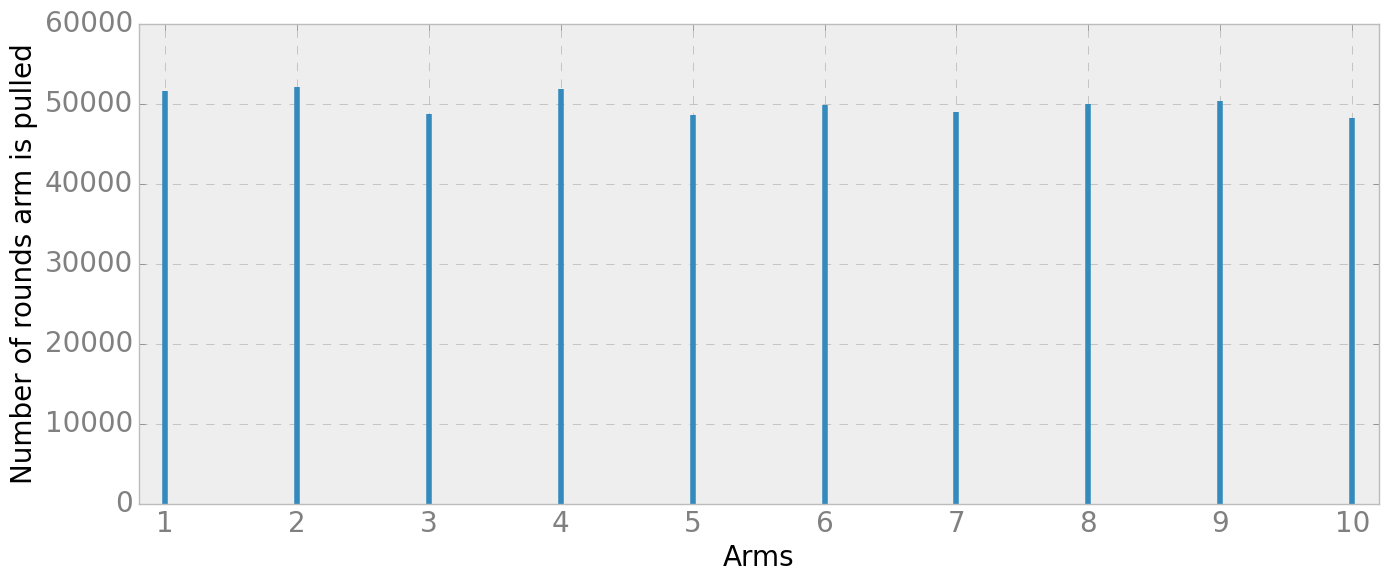}
     \vspace{-7mm}
     \caption{ The number of rounds each arm is pulled by the player when arms are using Strategy~\ref{Strategy}.}\label{ResTest}
     \vspace{-4mm}
\end{figure}

\vspace{-3mm}
\begin{table}[h]
  \begin{center}
    \caption{Summary of the numerical results.}
    \label{tab:table1}
    \begin{tabular}{c|c|c|r} 
      \textbf{T} &$\delta$& \textbf{Experimental total revenue} & \textbf{$\sqrt{KT\delta}$}\\
      \hline
      $5.10^5$& 2778 & 105169 & 117838\\
    \end{tabular}
  \end{center}
  \vspace{-4mm}
\end{table}

Fig.~\ref{ResTest} confirms our claim that by following Strategy\ref{Strategy} against a low-regret algorithm, arms will reach an equilibrium. This is evident from the balanced number of rounds the arms are pulled, in contrast to the ordinary MAB setting where the algorithm tends to favor choosing the best arm much more frequently than the others. Table~\ref{tab:table1} further supports our second claim that this strategy leaves the player with a cumulative reward less than $O(\sqrt{KT\delta})$.
\section{Conclusion}
In scenarios involving repeated interactions, converting a single-step collusion scenario into an equilibrium within the cumulative game necessitates the ability of each participant to identify instances where others deviate from the collusive behavior. This study illustrates that even when not all historical information is publicly accessible, the arms can implement a communication strategy enabling each of them to detect deviations, whether they involve falsifying player reports or manipulating shared information. The established equilibrium surpasses any low-regret MAB algorithm, resulting in reduced player revenues. Future research can focus on developing mechanisms that encompass not only traditional sequential learning in the classical MAB style but also integrate incentive mechanisms to effectively address the challenges highlighted in this paper.

\balance 
\bibliographystyle{unsrt} %plain
\bibliography{library}

\begin{thebibliography}{10}

\bibitem{principal-agentProblem}
Sylvain Chassang.
\newblock Calibrated incentive contracts.
\newblock {\em Econometrica}, 81(5):1935--1971, September 2013.

\bibitem{principal-agentProblem1}
Jean-Jacques Laffont and David Martimort.
\newblock The theory of incentives: The principal-agent model.
\newblock pages i--vi, 2002.

\bibitem{amin}
Kareem Amin, Afshin Rostamizadeh, and Umar Syed.
\newblock Learning prices for repeated auctions with strategic buyers.
\newblock {\em Advances in Neural Information Processing Systems}, 11 2013.

\bibitem{SideOb}
Shie Mannor and Ohad Shamir.
\newblock From bandits to experts: On the value of side-observations.
\newblock In {\em Proceedings of the 24th International Conference on Neural Information Processing Systems}, NIPS'11, page 684–692, Red Hook, NY, USA, 2011. Curran Associates Inc.

\bibitem{SideOb1}
St{\'{e}}phane Caron, Branislav Kveton, Marc Lelarge, and Smriti Bhagat.
\newblock Leveraging side observations in stochastic bandits.
\newblock In {\em Proceedings of the Twenty-Eighth Conference on Uncertainty in Artificial Intelligence, Catalina Island, CA, USA, August 14-18, 2012}, pages 142--151. {AUAI} Press, 2012.

\bibitem{SideOb2}
Swapna Buccapatnam, Atilla Eryilmaz, and Ness~B. Shroff.
\newblock Stochastic bandits with side observations on networks.
\newblock In {\em Measurement and Modeling of Computer Systems}, 2014.

\bibitem{pmlr-v99-braverman19b}
Mark Braverman, Jieming Mao, Jon Schneider, and S.~Matthew Weinberg.
\newblock Multi-armed bandit problems with strategic arms.
\newblock In Alina Beygelzimer and Daniel Hsu, editors, {\em Proceedings of the Thirty-Second Conference on Learning Theory}, volume~99 of {\em Proceedings of Machine Learning Research}, pages 383--416. PMLR, 25--28 Jun 2019.

\bibitem{ProfBook}
Ali~H. Sayed.
\newblock Adaptation, learning, and optimization over networks.
\newblock {\em Foundations and Trends in Machine Learning}, 7(4-5):311--801, 2014.

\bibitem{profbook1}
Ali~H. Sayed.
\newblock Adaptive networks.
\newblock {\em Proceedings of the IEEE}, 102(4):460--497, 2014.

\bibitem{AuerExp3}
Peter Auer, Nicol\`{o} Cesa-Bianchi, Yoav Freund, and Robert~E. Schapire.
\newblock The nonstochastic multiarmed bandit problem.
\newblock {\em SIAM Journal on Computing}, 32(1):48--77, 2002.

\bibitem{yuan2015convergence}
Kun Yuan, Qing Ling, and Wotao Yin.
\newblock On the convergence of decentralized gradient descent.
\newblock {\em SIAM Journal on Optimization}, 26(3):1835--1854, 2016.

\bibitem{Kayaalp_2022}
Mert Kayaalp, Stefan Vlaski, and Ali Sayed.
\newblock Dif-{MAML}: Decentralized multi-agent meta-learning.
\newblock {\em {IEEE} Open Journal of Signal Processing}, 3:71--93, 2022.

\bibitem{bubeck2012regret}
Sébastien Bubeck and Nicolò Cesa-Bianchi.
\newblock Regret analysis of stochastic and nonstochastic multi-armed bandit problems.
\newblock {\em Foundations and Trends® in Machine Learning}, 5, 04 2012.

\bibitem{metropolis}
Nicholas {Metropolis}, Arianna~W. {Rosenbluth}, Marshall~N. {Rosenbluth}, Augusta~H. {Teller}, and Edward {Teller}.
\newblock {Equation of State Calculations by Fast Computing Machines}.
\newblock {\em Journal of Chemical Physics}, 21(6):1087--1092, June 1953.

\end{thebibliography}

\end{document}